\documentclass[10pt,twocolumn,letterpaper]{article}

\usepackage[pagenumbers]{cvpr} 

\usepackage{algorithm}
\usepackage{algorithmic}
\usepackage{booktabs}
\usepackage[utf8]{inputenc} 
\usepackage{amsfonts}       
\usepackage{nicefrac}       
\usepackage{microtype}      
\usepackage{xcolor}         
\usepackage{mathtools}
\usepackage{multirow}
\usepackage{colortbl}
\usepackage{adjustbox}
\usepackage{pifont}
\definecolor{cvprblue}{rgb}{0.21,0.49,0.74}
\definecolor{maroonx}{RGB}{195,18,48}
\usepackage{dsfont}
\usepackage{enumitem}
\usepackage{subcaption}
\usepackage{bbold}
\definecolor{lightpastelpurple}{rgb}{0.69, 0.61, 0.85}
\colorlet{LightGreen}{lightpastelpurple!40}
\usepackage{makecell}

\definecolor{cvprblue}{rgb}{0.21,0.49,0.74}
\usepackage[pagebackref,breaklinks,colorlinks,allcolors=cvprblue]{hyperref}

\def\paperID{8295} 
\def\confName{CVPR}
\def\confYear{2026}

\title{Adaptive Cache Enhancement for Test-Time Adaptation of Vision-Language Models}

\author{Khanh-Binh Nguyen, Duc Thanh Nguyen\\
Deakin University\\
\and
Phuoc-Nguyen Bui, Hyunseung Choo\\
Sungkyunkwan University\\
}

\begin{document}
\maketitle
\begin{abstract}
Vision-language models (VLMs) demonstrate strong zero-shot generalization but degrade under distribution shifts in unlabeled downstream tasks. Test-Time Adaptation (TTA) enables online optimization during inference without annotations. Cache-based TTA methods maintain dynamic caches of low-entropy or high-confidence samples for efficient out-of-distribution adaptation. However, they encounter two key issues: (1) unreliable confidence metrics causing cache errors and reduced performance; and (2) inflexible decision boundaries failing to handle distributional variations. To address these, we propose the Adaptive Cache Enhancement (ACE) framework, which builds a robust cache by selectively storing high-confidence or low-entropy embeddings per class. It employs dynamic, class-specific thresholds initialized from zero-shot statistics and refined via exponential moving average and exploration-augmented updates, enabling adaptive class-wise boundaries for accurate predictions across diverse distributions.
Experiments on 15 benchmarks show ACE achieves state-of-the-art results, outperforming existing TTA methods in challenging out-of-distribution scenarios.

\end{abstract}    
\section{Introduction}
Vision-language models (VLMs) such as CLIP~\cite{CLIP} and ALIGN~\cite{jia2021scaling} exhibit remarkable zero-shot abilities in a wide range of tasks, including classification, retrieval, and segmentation~\cite{CoOp,MaPLe,raclip_retrieval,reco_retrieval,rpn_seg,seg2}. These models are trained to learn the semantic relationship between visual and textual information and can be used for image-text matching during inference. However, their effectiveness often decreases when handling unlabeled test data from domains that substantially differ from their training distribution~\cite{vlms1,vlms2,vlms3}.

\begin{figure}[!t]
    \centering
    \includegraphics[width=\columnwidth]{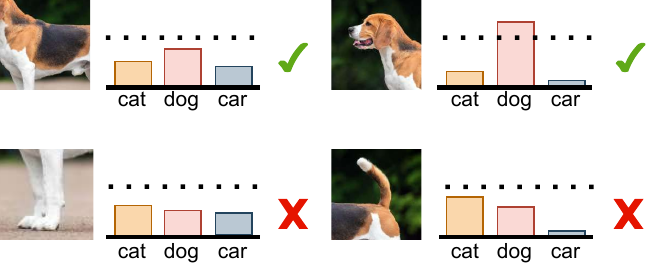}
    \caption{Noisy CLIP predictions on different views of a ``dog``. CLIP can be overconfident on the simple views and fail on others.}
    \label{fig:intro}
\vspace{-10pt}
\end{figure}

To address the challenge of performance decline of VLMs to distribution shifts, test-time adaptation (TTA) has emerged as a strong and effective strategy for improving their adaptability in out-of-distribution (OOD) cases. Unlike supervised fine-tuning or prompt learning, which requires expensive labeled data, TTA exploits unlabeled test data to dynamically adjust model predictions during inference. This characteristic makes TTA suitable for real-world applications, such as autonomous systems and medical imaging, where annotated datasets are often limited. 

Prompt-based TTA methods~\cite{tpt,difftpt,promptalign} reduce the prediction entropy of VLMs by optimizing domain-specific prompts. However, their prompting optimization requires repeated backpropagation through the encoders of VLMs, resulting in substantial computational cost and hence limiting their scalability in resource-limited settings. Cache-based TTA strategies~\cite{tda,boostadapter,dmn,dpe} dynamically assemble a cache of high-confidence samples guided by VLM prediction entropy metrics, thus fine-tuning predictions with minimal parameter changes. For example, TDA~\cite{tda} uses adaptive caching to enhance the calibration of predictions, while DMN~\cite{dmn} utilizes short- and long-term memory banks to leverage past insights. Similarly, DPE~\cite{dpe} improves multimodal alignment through dual-modal residual learning. 

However, as demonstrated in Figure~\ref{fig:intro}, CLIP predictions vary significantly under different augmentations, making the entropy-based sample selection unreliable and potentially hindering the adaptability of the model~\cite{entropy_unreliable,zero}. Furthermore, the decision boundaries of the VLMs in target domains become inconsistent. Even with TTA, these models frequently fail to adjust the decision boundaries to accommodate domain shifts, resulting in fluctuating predictions. Although several methods~\cite{dmn,dpe} attempt to overcome this issue by using mean textual prototypes from enriched prompts, they are limited in handling the variability from intra-class diversity shifts. 

To address the above challenges, we propose the Adaptive Cache Enhancement (ACE) framework, which constructs a robust cache by selectively storing high-confidence or low-entropy image embeddings for each class, directed by dynamic, class-specific thresholds initialized from zero-shot statistics and iteratively refined with an exponential moving average and exploration-augmented updates. By excluding overconfident samples from the cache, our method avoids overfitting, allows for adaptable class-specific decision boundaries, and ensures reliable and precise predictions across a broad range of data distributions. To this end, we make the following contributions in our work.
\begin{itemize}
\item We propose ACE, a novel framework for boosting up the performance of VLMs in domain shifts.

\item We introduce an adaptive and class-wise thresholding mechanism to enhance the reliability of predictions across diverse data distributions. Our thresholding mechanism is initialized from zero-shot statistics and refined via exponential moving average and exploration-augmented updates.

\item We validate the robustness and generality of our method through extensive experiments on 15 benchmark datasets and show the superiority of our ACE over current research in challenging OOD scenarios.
\end{itemize}

\section{Related Work}
\subsection{Vision-Language Models (VLMs)}
Large-scale pre-trained VLMs such as CLIP~\cite{CLIP} have demonstrated substantial proficiency in acquiring transferable representations that span both imagery and text. Extensive research has been conducted to optimize these models for downstream applications. Initial efforts, including prompt learning strategies~\cite{CoOp,cocoop}, achieve notable success by implementing minimal supervision in few-shot learning environments. Recent methods, e.g., Tip-Adapter~\cite{tipadapter} and TaskRes~\cite{taskres}, have effectively utilized feature memories derived from a limited number of labeled instances for rapid adaptation. These memory-based approaches have gained popularity due to their ability to facilitate quick domain adaptation with minimal computational burden, eliminating the requirement for full model fine-tuning. Nevertheless, their dependency on labeled target-domain data limits their applicability in real-world contexts, where such data are often not available. Our research focuses on test-time adaptation, which compels VLMs to acclimate to new domains during the inference stage without access to labeled data. This scenario closely reflects the real-world setting, advancing the development of flexible vision-language systems.

\subsection{Test-Time Adaptation (TTA) for VMLs}
Test-time adaptation (TTA) improves vision-language models (VLMs) adaptability to unfamiliar domains using only unlabeled test data~\cite{tpt,ma2024swapprompt,zero,rlcf}. Methods fall into prompt-based and cache-based categories. Prompt-based approaches refine textual embeddings during inference, such as TPT~\cite{tpt} for augmented view consistency and DiffTPT~\cite{difftpt} for diffusion-augmented diversity. In contrast, cache-based methods leverage historical test samples, including TDA~\cite{tda} for positive/negative feature caches, DMN~\cite{dmn} for static/dynamic memory integration, and DPE~\cite{dpe} for parallel prototype refinement. While prompt-based methods are simple, cache-based ones offer better adaptability and efficiency via lightweight domain modeling. Recent advances include BCA~\cite{zhou2025bayesian} (Bayesian priors with overhead), FreeTTA~\cite{dai2025free} (training-free online EM/GMM), COSMIC~\cite{huang2025cosmic} (semantic cliques), and AWT~\cite{zhu2024awt} (augmentation-weighting-transport), yet they neglect class-specific threshold dynamics under shifts. Building on cache-based techniques, our work enhances cache quality through reliable selection and predictive strength via adaptive decision boundaries; ACE distinctively employs EMA-refined thresholding to boost reliability and intra-class diversity handling.

\begin{figure*}[ht]
    \centering
    \includegraphics[width=0.9\textwidth]{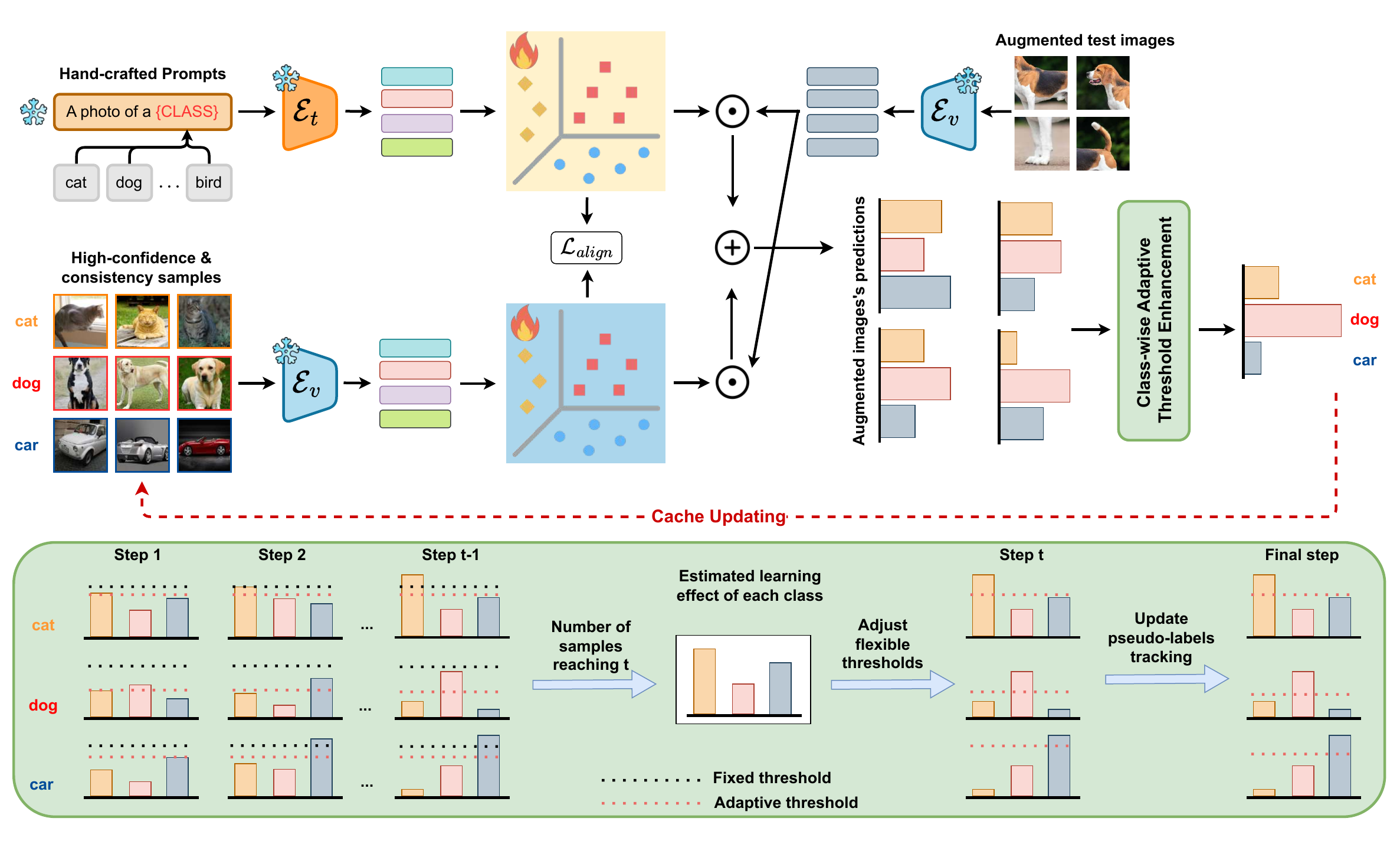}
    \caption{ \textbf{Overview of the ACE Method.} We introduce the Class-wise Adaptive Threshold Enhancement module, which refines the threshold online to allow more correct samples but low-confidence samples to be cached. By simultaneously minimizing entropy loss for the prototype residuals and creating highly reliable caches, ACE enhances the overall multimodal generalization and robustness.}
    \label{fig:main}
\vspace{-10pt}
\end{figure*}

\begin{table}[ht]
  \centering
  \setlength{\tabcolsep}{1.4mm}
    \centering
    \resizebox{\linewidth}{!}{%
    \begin{tabular}{lllll}
      \toprule
        Method & Key Mechanism & \thead{Adaptive\\ Thresholds} & \thead{Class\\ Specific} & \thead{Efficiency\\ Focus} \\
        \midrule
        BCA & Bayesian priors & No & Partial & No\\
        FreeTTA & Online EM/GMM & No & No & Yes*\\
        COSMIC & Semantic cliques & No & No & Yes\\
        AWT & Augment/Weight/Transport & No & No & Partial\\
        \midrule
        ACE (Ours) & \thead{Curriculum\\ + Exploration Thresholding} & \textbf{Yes} & \textbf{Yes} & \textbf{Yes}\\
      \bottomrule
    \end{tabular}%
    }
\caption{Methodology comparison. '*' denotes training-free.}
\vspace{-10pt}
\end{table}
\section{Proposed Method}

We propose an effective and reliable TTA strategy that includes a cache data structure, prototype residual learning modules, a flexible class-wise thresholding mechanism, and an adaptive threshold refinement module based on the statistic of pseudo-labels. We realize our method with CLIP~\cite{CLIP}, a state-of-the-art VLM model, and DPE~\cite{dpe}, a prototype residual vector learning technique.

\subsection{Preliminaries}

\subsubsection{Zero-shot CLIP Prediction.} 
CLIP is pre-trained on a large collection of image-text pairs to align visual and textual information within an embedding space. CLIP can be trained by optimizing a contrastive loss that maximizes the cosine similarity between matching image-text pairs.
CLIP comprises a visual encoder $\mathcal{E}_v$ and a text encoder $\mathcal{E}_t$, which map images and text into a shared latent space $\mathbb{R}^d$.
At inference time, an image $x$ is classified by measuring the similarity between its visual representation $\mathbf{z} = \mathcal{E}_v(x)$ and textual embeddings $\textbf{t}^c = \mathcal{E}_t(\mathcal{P}^c)$, where $\mathcal{P}^c$ is a text prompt describing a class $c$, and $C$ is the total number of classes.
The zero-shot CLIP prediction probability is then given by:
\begin{equation}
    \label{eq:clip}
    p^c_{\text{CLIP}}=\frac{\exp \left(\mathbf{z} \cdot \textbf{t}^c / \tau\right)}{\sum_{j=1}^{C} \exp \left(\mathbf{z} \cdot \textbf{t}^j / \tau\right)} 
\end{equation}
where $\tau$ is the temperature parameter, typically set to 0.01.

\subsubsection{Cache-based TTA.} Inspired by the key-value storage framework introduced in TIP-Adapter~\cite{zhang2022tip}, several cache-based TTA approaches~\cite{tda,dpe,boostadapter} have been proposed, which dynamically store and update high-confidence feature-label pairs during inference.
The cache is organized as a priority queue, where the priority of each entry is determined by the entropy of CLIP's predictions.

Let $\mathbf{W}_{\mathrm{CLIP}}=\left[\textbf{t}^1, \textbf{t}^2, \ldots, \textbf{t}^C\right]^{\top}$ be a stack of text embeddings for all classes. Cache item is represented as:
\begin{equation}
    \left\{\mathbf{F}_{\text{cache}}, \mathbf{L}_p, \mathcal{H}\left(\mathbf{F}_{\text{cache}} \mathbf{W}_{\text{CLIP}}^{\top}\right)\right\},
\end{equation}
where $\mathbf{F}_{\text {cache }}$ denotes the stored image features, $\mathbf{L}_p$ is a one-hot pseudo-label predicted by CLIP, and $\mathcal{H}(p)$ is the entropy:
\begin{equation}
    \mathcal{H}(p) = \left(-\sum_{i=1}^{C} p_i \log p_i\right)
\end{equation}

As test samples arrive sequentially, new entries are inserted into cache slots corresponding to their pseudo-labels. When the cache exceeds its capacity, the entries with the highest entropy are evicted.
During inference, a test feature $\mathbf{z}$ retrieves related cache items, and the final prediction is calculated by combining the CLIP logits with those derived from the cache:
\begin{equation}
    \label{eq:cls_logits}
    \text{logits}_{\mathrm{cls}}(\mathbf{z}) = \mathbf{z}\mathbf{W}_\text{CLIP}^{\top} + \mathcal{F}(\mathbf{z}\mathbf{F}_{\text{cache}}^{\top})\mathbf{L}_p,
\end{equation}
where $\mathcal{F}(x) = \alpha \exp(-\beta (1 - x))$ is a modulation function with a scaling factor $\alpha$ and a sharpness parameter $\beta$.

Following the DPE approach~\citet{dpe}, the feature for each cache slot, $\mathbf{F}_{\text{cache}}$, is computed as the mean of all stored features for that class, forming increasingly compact and representative class-specific prototypes as more data accumulates. The cache item is then constructed as:
\begin{equation}
    \left\{\mathbf{F}_{\text{cache}}, \mathbf{L}_p, \mathcal{H}\left(\mathbf{F}_{\text{cache}} \mathbf{W}_{\text{CLIP}}^{\top}\right) \leq \mathcal{T}\right\},
\end{equation}
where $\mathcal{T}$ is an entropy threshold for the caching decision.

\subsection{Adaptive Cache Enhancement (ACE)}
Cache-based strategies demonstrate notable effectiveness in various domains. However, they are grappling with two substantial issues. First, entropy is not a reliable indicator when the data distribution changes. Second, fixed decision boundaries are not adaptive to distributional changes. To address these challenges, we introduce two specialized modules that specifically address each issue while synergistically strengthening each other, as illustrated in Figure \ref{fig:main}.
In our framework, we explore two variants of ACE to accommodate different confidence estimation strategies: ACE-Entropy and ACE-Probability. ACE-Entropy employs entropy $\mathcal{H}(p)$ as the primary metric for cache selection and threshold refinement. In contrast, ACE-Probability uses maximum class probability $\max(p_c)$ for thresholding. Both is initialized from zero-shot statistics and updated via EMA, and both maintain computational parity, with no additional overhead.

\subsubsection{Curriculum Thresholding}
The task of dynamically setting class-specific thresholds in alignment with the evolving learning progress of the model is challenging. Ideally, the threshold for each class can be set specifically based on evaluation metrics, e.g., recognition accuracy or entropy as \(\mathcal{T}_t(c) = m_t(c) \cdot \mathcal{T}_0(c)\), where $\mathcal{T}_t(c)$ is the adaptive threshold for class $c$ at time step $t$, and $m_t(c)$ denotes the corresponding metric, whether it is accuracy or entropy. This ensures that classes with poorer performance, indicated by lower accuracy rates, are designated smaller thresholds (or conversely, higher entropy leads to larger thresholds), thereby permitting a greater number of their instances to be retained in the cache during the testing phase. However, dynamically adjusting these thresholds throughout the training process requires continuous accuracy estimates at time increments $t$, which impedes training efficiency.

In this study, we introduce Curriculum Thresholding, a thresholding mechanism that evaluates the learning status of classes without introducing any additional inference steps and/or requiring additional annotations. Central to our hypothesis is the postulation that when subjected to a high threshold, the quantity of confidently classified predictions ascribed to a particular class serves as a robust indicator of the learning progress for that class. Specifically, we proceed with the following steps:

{\small
\begin{align*}
    \setlength{\abovedisplayskip}{1.5pt}
    \setlength{\belowdisplayskip}{2.0pt}
\label{eqn:sigma}
\sigma_{t} (c)
    &= \sum_{n=1}^N \mathds{1}\left(\max\left(p(y|x)\right) \geq
    \mathcal{T}_{t-1}(c)\right) \cdot L_p^c,\\
    \text{or} &= \sum_{n=1}^N \mathds{1}\left(\min\left(e(y|x)\right) \leq
    \mathcal{T}_{t-1}(c)\right) \cdot L_p^c,\\
    L_p^c&=
    \begin{cases}
      1, & \text{if}\ \arg\max(p(y|x)) = c \\
      0, & \text{otherwise}
    \end{cases}
\end{align*}
}%
where $\sigma_t(c)$ measures the number of samples confidently predicted as class $c$ at time $t$, and $p(y|x)$ and $e(y|x)$ are the probability and entropy of the model to sample $x$.
We then normalize $\sigma_t(c)$ as \(\sigma_t(c) \leftarrow \frac{\sigma_t(c)}{\max_{c} \sigma_t} \label{eqn:normed}\)

Intuitively, a higher $\sigma_t(c)$ indicates that the model has learned class $c$ more effectively. Next, we compute an adaptive threshold $\mathcal{T}_t(c)$ for class $c$ at time step $t$ using the Exponential Moving Average (EMA) as,
\begin{equation}
\label{eqn:basemap}
\mathcal{T}_t(c)=\delta \mathcal{T}_{t-1}(c) + (1 - \delta) \cdot m_{t}(c)
\end{equation}
where $m_t(c) = \sigma_t(c) \cdot m_{t-1}(c)$ is the latest performance metric (accuracy or entropy) and $\delta$ is an EMA factor.

Nevertheless, considering that Eq.~(\ref{eqn:basemap}) is executed solely during the testing phase, it should be anticipated that initial performance might not be optimal. To improve the adaptability and reduce overfitting of the model, particularly for classes well-represented in the dataset, the initial threshold at time $t=0$ is determined using statistics from zero-shot learning. This statistic is simply the average of the entropy and probability across all samples in the test set. This initialization effectively captures the true behavior of both the model and the test set, ensuring reliable and meaningful results. It is important to highlight that incorporating CT is essentially cost-effective. In practice, whenever the confidence level of a sample's prediction surpasses the threshold $\mathcal{T}$, the sample, along with its predicted class, is flagged. Flagged instances will be retained in the cache in subsequent iterations.

\subsubsection{Online Thresholding Adaptation}
The CLIP model exhibits limitations in terms of its applicability across all datasets. Consequently, in scenarios such as zero-shot evaluations or even during TTA, predictions may not always be made for certain classes. It is critical to monitor classes that have been encountered infrequently or not at all. For these infrequently seen or never-seen classes, we implement a strategy for more intensive threshold exploration. This approach enhances the diversity of the cache and accelerates the adaptation to classes that exhibit a long-tailed distribution. To streamline our analysis, we categorize any class with fewer than 10 predictions as rarely seen classes.
{\small
\begin{align}
\label{eqn:adapt}
    \nonumber \text{Never seen}:
    \mathcal{T}_{t}(c) 
    & = \sum_1^{C}\mathds{1}\left(M_{c,t}= 0\right) * \mathcal{T}_{t}(c) * (1 - \gamma) \\
    \nonumber \text{Rarely seen}:
    \mathcal{T}_{t}(c)
    & = \sum_1^{C}\mathds{1}\left(M_{c,t} \leq 10\right) * \mathcal{T}_{t}(c) * (1 - \gamma * 0.5)
\end{align}
}%
where $\gamma$ is an adaptation rate and $M_{c,t}$ is the number of samples of class $c$ at time $t$ in the cache.

The online thresholding adaption ensures that the best-learned class reaches the full threshold $\mathcal{T}$, while the less-learned classes get proportionally lower thresholds, encouraging more of their samples to be trained. As the training progresses, the model improves and the thresholds converge towards $\mathcal{T}$. Notably, the thresholds are not guaranteed to increase over time; they may decrease if predictions become less confident in later iterations.

\begin{table*}[t]
\renewcommand\arraystretch{1.2}
\renewcommand{\tabcolsep}{3pt}
  \centering
  \resizebox{0.9\linewidth}{!}{
  \begin{tabular}{lccccccc}
    \toprule
    {Method}      & ImageNet   &  ImageNet-A  &  ImageNet-V2  & ImageNet-R 
    & ImageNet-S   
    &{Average}  & {OOD Average}     \\
  \midrule
  CLIP-ResNet-50~\cite{radford2021learning}       &58.16&	21.83&	51.41	&56.15	&33.37&	44.18&	40.69           \\ 
  \midrule
  Ensemble          &  59.81&	23.24&	52.91	&{60.72}	&35.48&	46.43&	43.09  \\
  CoOp~\cite{zhou2022learning}            &  63.33 &  23.06  &   55.40     &  56.60   &  34.67   &     46.61   &     42.43       \\
  \midrule
  TPT~\cite{shu2022test}          &  60.74 & 26.67  &    54.70   &  59.11   &  35.09   &    47.26    &     43.89    \\
  DiffTPT~\cite{feng2023diverse} & 60.80 & 31.06 & 55.80 & 58.80 & 37.10 & 48.71 & 45.69 \\
  TDA~\cite{karmanov2024efficient} & 61.35 & 30.29 & 55.54 & 62.58 & 38.12 & 49.58 & 46.63 \\ 
  TPS~\cite{sui2024just}	& 61.47 & 30.48 &	54.96 &	62.87 &	37.14&	49.38&	46.36 \\
  DMN-ZS~\cite{zhang2024dual}	& 63.87 &	28.57 &	56.12&	61.44 &	39.84 &	49.97 &	46.49\\ 
  DPE~\cite{dpe} & 63.41 & 30.15 & 56.72 & 63.72 & 40.03 & 50.81  & 47.66 \\
  CRG~\cite{zhai2025mitigatingcachenoisetesttime} & \textbf{65.26} & 29.69 & 56.07 & - & - & - & - \\
  \rowcolor{cvprblue!20}
  \textbf{ACE-Probability} & \underline{64.41} & \underline{33.50} & \underline{57.50} & \textbf{63.95} & \underline{40.15} & \underline{51.75}  & \underline{49.37} \\
  \rowcolor{cvprblue!20}
  \textbf{ACE-Entropy} & 64.30 & \textbf{33.55} & \textbf{57.51} & \underline{63.82} & \textbf{40.25} & \textbf{51.78}  & \textbf{49.38} \\
  \midrule
  \midrule
  CLIP-ViT-B/16~\cite{radford2021learning}       & 66.73 &	47.87&	60.86&	73.98&	46.09&	59.11&	57.20             \\
  \midrule
  Ensemble          &  68.34&	49.89&	61.88&	{77.65}&	48.24&	61.20&	59.42  \\
  CoOp~\cite{zhou2022learning}          &  71.51 &  49.71  &   64.20     &  75.21   &  47.99   &   61.72     &   59.28  \\
  \midrule
  TPT~\cite{shu2022test}           &  68.98  & 54.77  & 63.45       &  77.06   &  47.94   &   62.44     &   60.81 \\
  DiffTPT~\cite{feng2023diverse}           &  70.30  & 55.68  & 65.10       &  75.00   &  46.80   &   62.28     &   60.52 \\
  TDA~\cite{karmanov2024efficient} & 69.51 & 60.11 & 64.67 & 80.24 & 50.54 & 65.01 & 63.89 \\ 
  TPS~\cite{sui2024just}	&70.19&	60.08&	64.73&	80.27 &	49.95&	65.04&	63.76 \\
  DMN-ZS~\cite{zhang2024dual}	& 72.25 &	58.28 &	65.17 &	78.55 &	\textbf{53.20} &	65.49 &	63.80\\
  DPE~\cite{dpe} & 71.91 & 59.63 & 65.44 & 80.40 & 52.26 & 65.93 & 64.43 \\
  CRG~\cite{zhai2025mitigatingcachenoisetesttime} & \textbf{75.01} & \textbf{63.67} & 64.66  & - & - & - & - \\
  \rowcolor{cvprblue!20}
  \textbf{ACE-Probability} & 72.56 & \underline{63.54} & \underline{65.86} & \underline{81.01} & 52.60 & \underline{67.05} & \underline{66.01} \\
  \rowcolor{cvprblue!20}
  \textbf{ACE-Entropy} & \underline{72.57} & 63.47 & \textbf{65.91} & \textbf{81.09} & \underline{52.66} & \textbf{67.10} & \textbf{66.05} \\
    \bottomrule
  \end{tabular}
    }
\caption{Performance comparisons on robustness to natural distribution shifts. We present top-1 accuracy (\%) results for all evaluated methods employing both ResNet-50 and ViT-B/16 visual backbones of CLIP. The best and second-best results are highlighted in \textbf{bold} and \underline{underlined}, respectively.}
\label{tab:ood-main}
\vspace{-10pt}
\end{table*}

\subsubsection{Test-Time Logit Inference and Loss Function.}
We adopt DPE, the residual learning method proposed by~\citet{zhang2024dual}, to define textual and visual embedding prototypes as $\mathbf{t}_c = \frac{1}{S} \sum_{i=1}^{S} \mathcal{E}_t(\mathcal{P}^c_i)$ and $\mathbf{v}_c = \frac{1}{S_c} \sum_{m=1}^{S_c} \mathbf{z}^c_m$. Here, $S$ is the number of prompts, $S_c \leq M$ is the total count of image features stored in the queue corresponding to class $c$, and $\mathbf{z}$ is the image features of $x$ that surpass the threshold $\mathcal{T}$. Once the two sets of multi-modal prototypes have been adjusted based on the last test sample, they are used to serve as the initial configuration for further refinement with the current test input. The preliminary outcomes derived from these steps are subsequently utilized to fine-tune the prototypes, making them adapted to $x$ as,
\begin{equation}
    \mathbf{t}_c \leftarrow \frac{\mathbf{t}_c + \hat{\mathbf{t}}_c}{\|\mathbf{t}_c + \hat{\mathbf{t}}_c\|}, \quad \mathbf{v}_c \leftarrow \frac{\mathbf{v}_c + \hat{\mathbf{v}}_c}{\|\mathbf{v}_c + \hat{\mathbf{v}}_c\|}.
\end{equation}
where $\hat{\mathbf{t}}_c \in \mathbb{R}^{C\times d}$ and $\hat{\mathbf{v}}_c \in \mathbb{R}^{C\times d}$ are the learnable residual parameters that are initialized as zero and are used to optimize the prototypes for each given test input $x$.
Similarly to \citet{zhang2024dual}, we optimize these residual parameters to promote consistent predictions in a total of $N$ different augmented views of each given test image $x$ using the objective of unsupervised entropy minimization,
{\small
\begin{align}
    \nonumber \mathcal{L}_{\mathtt{aug}} &= \mathcal{H}(\mathbb{P}_{\mathtt{ACE}}(x)) \\
    &= -\sum_{c=1}^C \mathbb{P}_{\mathtt{ACE}}(y=y_c | x) \log \mathbb{P}_{\mathtt{ACE}}(y=y_c | x)
\end{align}
}
where
{\small
\begin{align}
    \nonumber \mathbb{P}_{\mathtt{ACE}}(x) &= \frac{1}{\rho N} \sum_{n=1}^{N}\mathds{1}[\mathcal{H}\left(\mathbb{P}_{\mathtt{Proto}}(\mathcal{A}_n (x)\right) \leq \mathcal{T}_e] \,\mathbb{P}_{\mathtt{Proto}}(\mathcal{A}_n (x))\\
\nonumber \text{or} &= \frac{1}{\rho N} \sum_{n=1}^{N}\mathds{1}[p\left(\mathbb{P}_{\mathtt{Proto}}(\mathcal{A}_n (x)\right) \geq \mathcal{T}_p] \,\mathbb{P}_{\mathtt{Proto}}(\mathcal{A}_n (x))
\end{align}
}
where $\mathcal{A}_n(x)$ is an augmented view of $x$, $T_e$ and $T_p$ are the entropy threshold and probability threshold, respectively.

We adopt the alignment loss in DPE:
{\small
\begin{equation}
\label{eq:alignment}
     \mathcal{L}_{\mathtt{align}} = \frac{1}{C}\sum_{c=1}^C - \log \left(\frac{\exp (\textbf{t}_{c}^{\,\,\top} \textbf{v}_{c})}{\sum\nolimits_{c'} \exp (\textbf{t}_{c}^{\,\,\top} \textbf{v}_{c'})} \frac{\exp (\textbf{t}_{c}^{\,\,\top} \textbf{v}_{c})}{\sum\nolimits_{c'} \exp (\textbf{t}_{c'}^{\,\,\top} \textbf{v}_{c})}\right)
\end{equation}
}%

The final objective for optimizing the multi-modal prototypes $\mathbf{t}, \mathbf{v}$ is defined as,
\begin{equation}
\label{eq:update}
    \hat{\mathbf{t}}, \hat{\mathbf{v}} = \arg\min_{\mathbf{t,v}}\{\mathcal{L}_{\mathtt{aug}} + \lambda \mathcal{L}_{\mathtt{align}}\},
\end{equation}

After optimizing the prototypes for each test sample $x$, we evolve the online textual prototypes $\mathbf{t}$, and update the priority queues to re-compute the visual prototypes $\mathbf{v}$. Then, we perform the testing as usual as in Eq.~(\ref{eq:cls_logits}).

\begin{table*}[t]
\renewcommand\arraystretch{1.3}
\renewcommand{\tabcolsep}{2pt}
  \vspace{0pt}
  \centering
    \begin{tabular}{lp{1.2cm}<{\centering}p{1.2cm}<{\centering}p{1.2cm}<{\centering}p{1.2cm}<{\centering}p{1.2cm}<{\centering}p{1.2cm}<{\centering}p{1.2cm}<{\centering}p{1.2cm}<{\centering}p{1.2cm}<{\centering}p{1.2cm}<{\centering}p{1.2cm}<{\centering}}
      \toprule
      Method & Aircraft & Caltech & Cars & DTD & EuroSAT & Flower & Food101 & Pets & SUN397 & UCF101 & Average \\
      \midrule
      CLIP-ResNet-50 & 15.66 & 85.88 & 55.70 & 40.37 & 23.69 & 61.75 & 73.97 & 83.57 & 58.80 & 58.84 & 55.82 \\
      \midrule
      CoOp~\cite{CoOp} & 15.12 & 86.53 & 55.32 & 37.29 & 26.20 & 61.55 & 75.59 & 87.00 & 58.15 & 59.05 & 56.18 \\
      \midrule
      TPT~\cite{tpt}  & 17.58 & 87.02 & 58.46 & 40.84 & 28.33 & 62.69 & 74.88 & 84.49 & 61.46 & 60.82 & 57.66 \\
      DiffTPT~\cite{Feng2023DiverseDA} & 17.60 & 86.89 & \textbf{60.71} & 40.72 & 41.04 & 63.53 & \textbf{79.21} & 83.40 & 62.72 & 62.67 & 59.85 \\
      TDA~\cite{tda}  & 17.61 & 89.70 & 57.78 & 43.74 & 42.11 & 68.74 & 77.75 & 86.18 & 62.53 & \textbf{64.18}  & 61.03\\ 
      DPE~\cite{dpe}  & 19.80 & 90.83 & 59.26 & 50.18 & 41.67 & 67.60 & \underline{77.83} & 85.97 & 64.23 & 61.98  & 61.93\\ 
      CRG~\cite{zhai2025mitigatingcachenoisetesttime} & 18.09 & 90.12 & 57.92 & 51.89 & 46.80 & \textbf{71.09} & 75.76 & 85.91 & 63.11 & \underline{63.58} & 62.42 \\
      \rowcolor{cvprblue!20}
      \textbf{ACE-Probability} & \underline{21.33}  &  \underline{90.91}  & 59.84  & \textbf{55.20} & \textbf{51.02} & \underline{70.28} & 76.83 & \textbf{87.95} & \textbf{65.46} & 61.83 & \textbf{64.07} \\ 
      \rowcolor{cvprblue!20}
      \textbf{ACE-Entropy} & \textbf{21.48}  & \textbf{90.99}  & \underline{60.00}  & \underline{53.90} & \underline{48.54} & 70.08 & 76.91 & \underline{87.54} & \underline{65.38} & 62.60 & \underline{63.74} \\ 
      \midrule
      \midrule
      CLIP-ViT-B/16 & 23.67 & 93.35 & 65.48 & 44.27 & 42.01 & 67.44 & 83.65 & 88.25 & 62.59 & 65.13 & 63.58 \\
      \midrule
      CoOp~\cite{CoOp} & 18.47 & 93.70 & 64.51 & 41.92 & 46.39 & 68.71 & 85.30 & 89.14 & 64.15 & 66.55 & 63.88 \\
      \midrule
      TPT~\cite{tpt}  & 24.78 & 94.16 & 66.87 & 47.75 & 42.44 & 68.98 & 84.67 & 87.79 & 65.50 & 68.04 & 65.10 \\
      DiffTPT~\cite{Feng2023DiverseDA} & 25.60 & 92.49 & 67.01 & 47.00 & 43.13 & 70.10 & \textbf{87.23} & 88.22 & 65.74 & 62.67 &65.47 \\
      TDA~\cite{tda} & 23.91 & 94.24 & 67.28 
      & 47.40 & \underline{58.00} & 71.42 & 86.14 & 88.63 & 67.62 & 70.66 & 67.53 \\ 
      DPE~\cite{dpe} & 28.95 & 94.81 & 67.31 
      & 54.20 & 55.79 & 75.07 & \underline{86.17} & \underline{91.14} & 70.07 & 70.44 & 69.40 \\ 
      CRG~\cite{zhai2025mitigatingcachenoisetesttime} & 26.58 & 93.57 & 66.89 & \underline{56.38} & \textbf{59.81} & 75.94 & 85.95 & 91.20 & 68.36 & 70.31 & 69.50 \\
      \rowcolor{cvprblue!20}
      \textbf{ACE-Probability} & \underline{29.46}  &  \underline{95.13}  & \textbf{68.72}  & \underline{56.38} & 54.28 & \textbf{77.47} & 85.88 & \textbf{91.47} & \underline{70.83} & \underline{73.62} & \textbf{70.32} \\ 
      \rowcolor{cvprblue!20}
      \textbf{ACE-Entropy} & \textbf{29.49}  &  \textbf{95.33}  & \underline{68.55}  & \textbf{56.56} & 53.63 & \underline{77.30} & 85.88 & \textbf{91.47} & \textbf{70.88} & \textbf{73.72} & \underline{70.28} \\ 
      \bottomrule
    \end{tabular}
  \caption{Performance comparisons on cross-dataset generalization across 10 fine-grained datasets. The best and second-best results are highlighted in \textbf{bold} and \underline{underlined}, respectively.}
  \label{tab:fine-grained}
  \vspace{-10pt}
\end{table*}
\section{Experiments}
In this section, we assess the robustness of the proposed ACE method to natural distribution shifts and its cross-dataset generalization across 15 different datasets. In addition, we compare the test-time efficiency of ACE against existing approaches. Lastly, we conduct ablation studies to examine the impact of various components and design choices.

\subsection{Experimental Setup} 
\paragraph{Datasets} Following prior research~\cite{tda,dpe}, we evaluate our method in two main benchmarking scenarios: cross-dataset generalization and robustness to natural distribution shifts. (1) For the cross-dataset generalization scenario, we carry out evaluations across 10 varied recognition datasets, including FGVC Aircraft~\cite{aircraft}, Caltech101~\cite{caltech101}, Stanford Cars~\cite{cars}, DTD~\cite{Dtd}, EuroSAT~\cite{helber2019eurosat}, Flowers102~\cite{oxfordflower}, Food101~\cite{bossard2014food}, Oxford Pets~\cite{oxford_pets}, SUN397~\cite{sun397}, and UCF101~\cite{ucf101}. These datasets collectively offer a comprehensive benchmark to assess the adaptability and generality of models across different datasets. (2) To evaluate the robustness of our method against natural distribution shifts, we test our method with ImageNet~\cite{deng2009imagenet} and its out-of-distribution variants, including ImageNet-A~\cite{hendrycks2021natural} and ImageNet-V2~\cite{recht2019imagenet}. This part of the evaluation assesses our method's robustness under varying distribution shifts. For datasets with particularly large test sets, e.g., Food101, ImageNet, SUN397, ImageNet-V2, and ImageNet-A, we sample 2,000 test instances for evaluation, adhering to the experimental protocol in DiffTPT~\cite{Feng2023DiverseDA}.

\paragraph{Implementation details}
Following previous studies~\cite{tda, dpe}, we select ResNet-50 and ViT-B/16 as visual encoders for CLIP, employing hand-crafted prompts as described in the supplementary material. In line with the method by~\cite{tpt}, we create 63 augmented views per test image. The AdamW optimizer is used with the learning rate set to 0.0005 for the residual parameters in one iteration. Hyperparameters are set as follows: $\alpha=6$, $\beta=5.0$, $\delta=0.95$, $\gamma=0.02$, $\lambda=0.5$, and queue size $M=16$ (since ACE shows resilience to noisy pseudo-labels and benefits from additional samples for class distribution modeling). All experiments are conducted on a single NVIDIA A100 GPU with 40GB of memory.

\subsection{Results}
\paragraph{Robustness to Natural Distribution Shifts.}
Our method reliably excels in adapting to varied image datasets with natural shifts, as highlighted in Table \ref{tab:ood-main}. 
Compared to TDA~\cite{tda} and DPE~\cite{zhang2024dual}, both our probability and entropy variants outperform them across 4 out of 5 datasets with notable differences, tested across ResNet-50 and ViT-B/16 backbones. Specifically, our ACE achieves top OOD average accuracies of 49.38\% with ResNet-50 and 66.05\% with ViT-B/16. While CRG~\cite{zhai2025mitigatingcachenoisetesttime} occasionally surpasses our method, we found this is due to the limited use of data augmentation in our experiments. Our method also outperforms the few-shot learning technique CoOp~\cite{CoOp}, underscoring our method’s robust ability to adapt VLMs to OOD scenarios.

\paragraph{Cross-Datasets Generalization.}
In Table~\ref{tab:fine-grained}, we evaluate the generality of our ACE-Probability and ACE-Entropy variants across 10 diverse fine-grained datasets, and compare them with state-of-the-art approaches. Due to significant distributional differences among these datasets, performance varies considerably. However, our methods achieve average accuracy improvements of 1.65\% and 0.82\% over the best-performing methods on the CLIP-ResNet-50 and ViT-B/16 backbones, respectively, surpassing competitors in 6 out of 10 datasets. Notably, on the challenging DTD dataset, characterized by complex texture categories, our approach consistently outperforms prior works across both CLIP backbones by a large margin. Additionally, our approach significantly outperforms the few-shot prompt learning method CoOp~\cite{CoOp} in all datasets. These findings highlight the generality and adaptability of our approach in test-time domain transfer, making it highly suitable for real-world applications.

\subsection{Ablation study}
\subsubsection{Cache size and zero-shot statistics.} 
Table~\ref{tab:ablation} presents an ablation study on the impact of zero-shot cache initialization and cache size on the performance of our ACE across five datasets. Table~\ref{tab:ablation} (a) compares DPE with our ACE in two settings: with and without zero-shot (ZS) cache. As shown in the results, ACE with ZS cache achieves higher average accuracies (95.33\% on Caltech and 77.30\% on Flower with ViT-B/16) compared to DPE, highlighting the benefit of ZS cache initialization. Table~\ref{tab:ablation} (b) examines the effect of cache size (M) using probabilistic (Prob.) and entropy (Ent.) quantities. The results show that larger cache sizes (e.g., M = 16) consistently improve the classification performance of CLIP (95.33\% on Caltech and 73.72\% on UCF with ViT-B/16). These achievements highlight ACE's ability in mitigating caching noise and enhancing classification accuracy of VLMs with increased cache capacity.

\begin{table}[ht]
  \centering
  \setlength{\tabcolsep}{1.4mm}
  \begin{subtable}[t]{\columnwidth}\small
    \centering
    \resizebox{\linewidth}{!}{%
    \begin{tabular}{llccccc}
      \toprule
      Setting & Strat. & Caltech & DTD & Flower & Pets & UCF \\
      \midrule
       DPE & - & 90.83 & 50.18 & 67.60 & 85.97 & 61.98\\ \midrule
       ACE w/o ZS cache & Prob. & 90.55 & 52.54 & 69.51 & 87.41 & 61.81 \\
       ACE w/ ZS cache & Prob.  & 90.91 & \textbf{55.20} & \textbf{70.28} & \textbf{87.95} & 61.83 \\\midrule
       ACE w/o ZS cache & Ent.  & 90.39 & 52.19 & 69.39 & 87.00 & 62.18 \\
       ACE w/ ZS cache & Ent.  & \textbf{90.99} & 53.90 & 70.08 & 87.54 & \textbf{62.60} \\
       \midrule
       \midrule
       DPE & -  & 94.81 & 54.20 & 75.07 & 91.14 & 70.44\\\midrule
       ACE w/o ZS cache & Prob.  & 95.06 & 56.21 & 75.80 & 91.21 & 72.06 \\
       ACE w/ ZS cache & Prob.  & 95.13 &56.38 & \textbf{77.47} & \textbf{91.47} & 73.62 \\\midrule
       ACE w/o ZS cache & Ent.  & 95.12 & 56.09 & 74.99 & 91.12 & 72.16 \\
       ACE w/ ZS cache & Ent.  & \textbf{95.33} & \textbf{56.56} & 77.30 & \textbf{91.47} & \textbf{73.72} \\
      \bottomrule
    \end{tabular}%
    }
    \caption{Impact of zero-shot cache}
  \end{subtable}\hfill
  \begin{subtable}[t]{\columnwidth}\small
    \centering
    \resizebox{\linewidth}{!}{%
    \begin{tabular}{llccccc}
      \toprule
      Strategy & Cache size & Caltech & DTD & Flower & Pets & UCF \\
      \midrule
      Prob. & M = 3 & 90.63 & 51.30 & 69.96 & 85.94 & 60.32 \\
      Ent. & M = 3 & 90.91 & 51.12 & 68.17 & 86.02 & 60.69 \\
      Prob. & M = 16 & 90.91 & \textbf{55.20} & \textbf{70.28} & \textbf{87.95} & 61.83 \\
      Ent. & M = 16 & \textbf{90.99} & 53.90 & 70.08 & 87.54 & \textbf{62.60} \\
      \midrule
      \midrule
      Prob. & M = 3 & 95.09 & 55.02 & 77.39 & 89.94 & 71.74 \\
      Ent. & M = 3 & 95.29 & 54.31 & \textbf{77.59} & 90.22 & 73.35 \\
      Prob. & M = 16 & 95.13 & 56.38 & 77.47 & \textbf{91.47} & 73.62 \\
      Ent. & M = 16 & \textbf{95.33} & \textbf{56.56} & 77.30 & \textbf{91.47} & \textbf{73.72} \\
      \bottomrule
    \end{tabular}%
    }
    \caption{Impact of cache size}
  \end{subtable}
\caption{Ablation study on the impact of zero-shot cache initialization and cache size toward ACE.
}
\label{tab:ablation}
\end{table}

\subsubsection{Visualization of cache features using t-SNE} 
\begin{figure}
    \centering
    \includegraphics[width=0.495\linewidth]{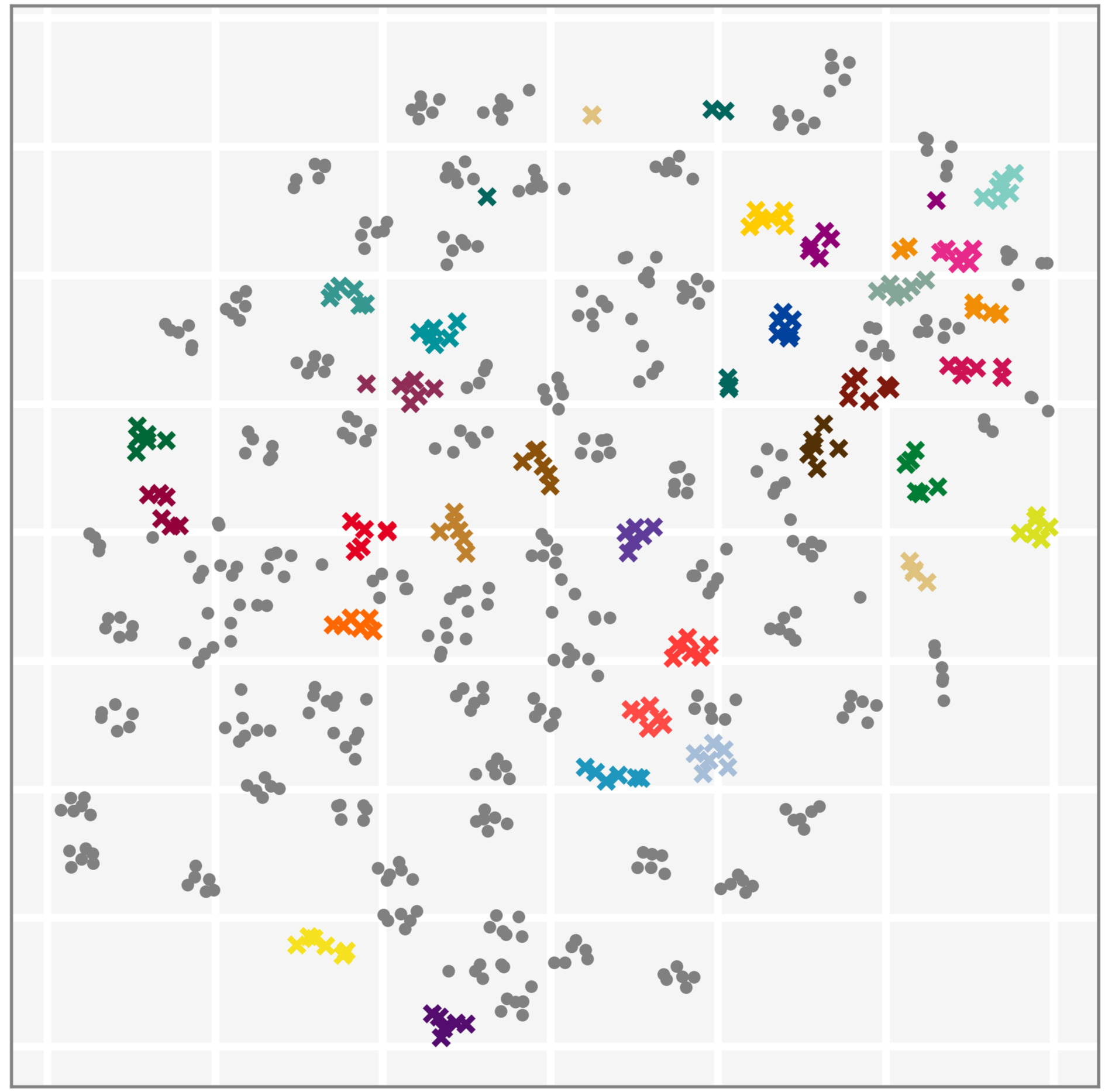}
    \includegraphics[width=0.495\linewidth]{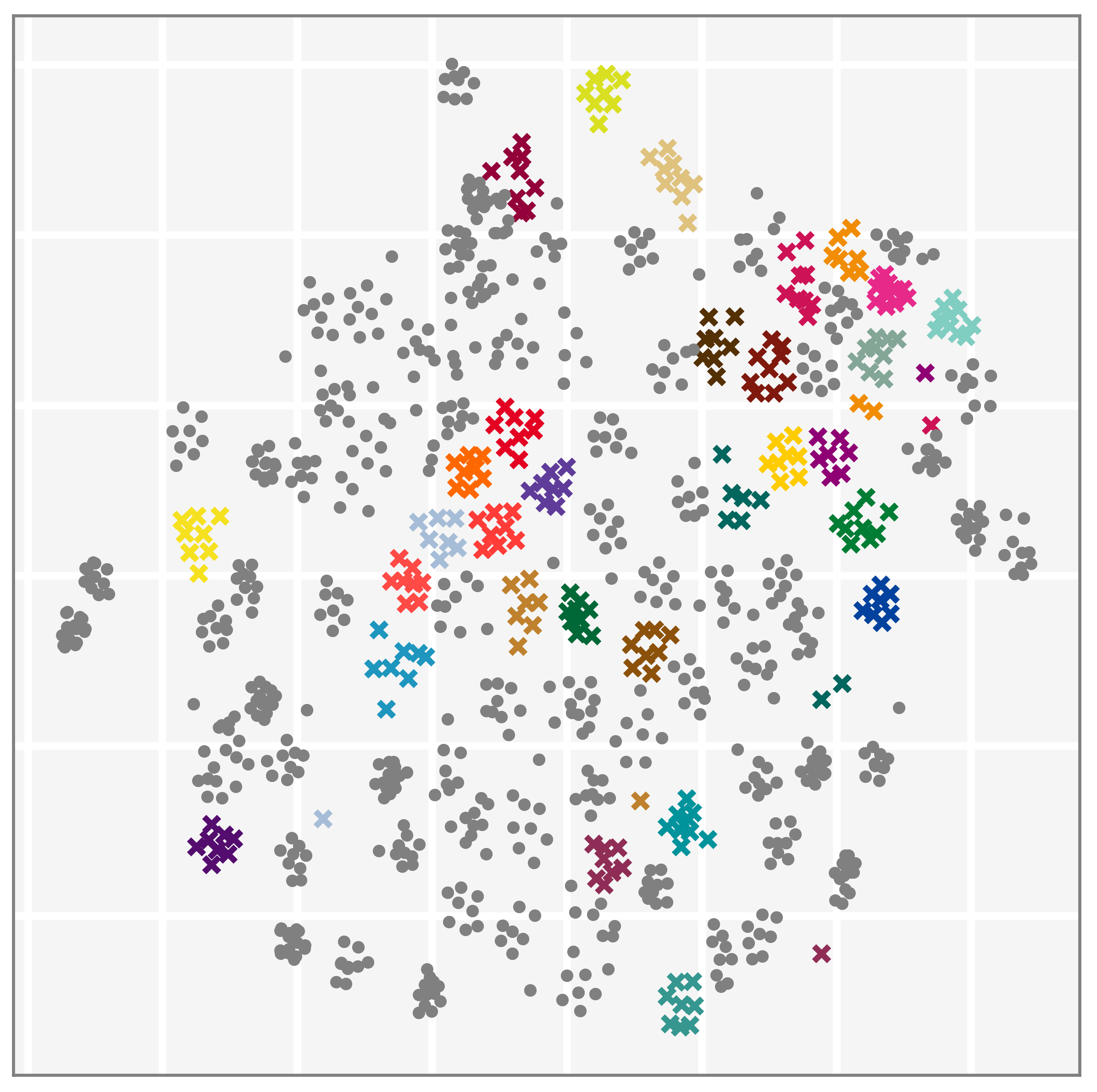}
    \caption{t-SNE~\cite{van2008visualizing} visualizations of the stored image features in the cache between DPE~\cite{zhang2024dual} and ACE-Entropy.}
    \label{fig:tsne}
    \vspace{-20pt}
\end{figure}

Figure~\ref{fig:tsne} displays t-SNE~\cite{van2008visualizing} of the cached image features (with cache size $M=6$) for both DPE~\cite{zhang2024dual} and our method on the Food101 \cite{bossard2014food} dataset. Features from 10\% of randomly selected classes are shown in different colors, while all others are indicated in gray. The visualizations demonstrate that our priority queue approach successfully gathers high-confidence samples, leading to increasingly representative visual prototypes over time.

\subsubsection{Efficiency Analysis}
The efficiency analysis in the Table \ref{table:efficiency} shows the trade-off of different methods. Baseline zero-shot CLIP achieves 59.81\% accuracy in just 9 minutes, serving as a reference for minimal overhead, while prompt-based approaches like TPT and DiffTPT deliver modest gains of +0.93\% and +0.99\% at significantly higher costs of 9 hours 15 minutes and 20 hours, respectively, due to backpropagation-intensive prompt optimization. Cache-based methods, including TDA (+1.54\% in 1 hour 5 minutes), TPS (+1.66\% in 55 minutes), and DPE (+3.60\% in 1 hour 50 minutes), offer better efficiency through lightweight mechanisms. Notably, the proposed ACE variants stand out, with ACE-Probability attaining the highest accuracy of 64.41\% (+4.60\% gain) and ACE-Entropy closely following at 64.30\% (+4.49\% gain), both in 2 hours 40 minutes; this moderate runtime increase reflects ACE's dynamic threshold refinement via exponential moving average and exploration updates, enabling superior robustness and generalization without prohibitive computational demands, positioning it as a state-of-the-art solution for practical out-of-distribution adaptation.

\begin{table}[t]
\resizebox{0.9\linewidth}{!}{
\begin{tabular}{lrcc}
\toprule
Method  & Testing Time & Accuracy & Gain \\ 
\midrule
CLIP~\cite{radford2021learning}  &  9 min &  59.81&  -\\
TPT~\cite{shu2022test} &   9 h 15 min &  60.74  & +0.93  \\
DiffTPT~\cite{feng2023diverse} &  20 h 00 min &  60.80  & +0.99  \\
TDA~\cite{karmanov2024efficient} &  1 h 5 min &  61.35  & +1.54  \\
TPS~\cite{sui2024just} & 55 min  &  61.47  & +1.66  \\
DPE &   1 h 50 min  & 63.41 & +3.60 \\
\rowcolor{cvprblue!20} \textbf{ACE-Probability} & 2 h 40 min & \textbf{64.41} & 4.60 \\
\rowcolor{cvprblue!20} ACE-Entropy & 2 h 40 min & 64.30 & 4.49 \\
\bottomrule
\end{tabular}
}
\caption{Efficiency analysis of ACE vs other methods.}
\label{table:efficiency}
\vspace{-15pt}
\end{table}

\subsubsection{Cache reliability} 
\begin{figure}[!t]
	\centering
        \centerline{\includegraphics[width=\linewidth]{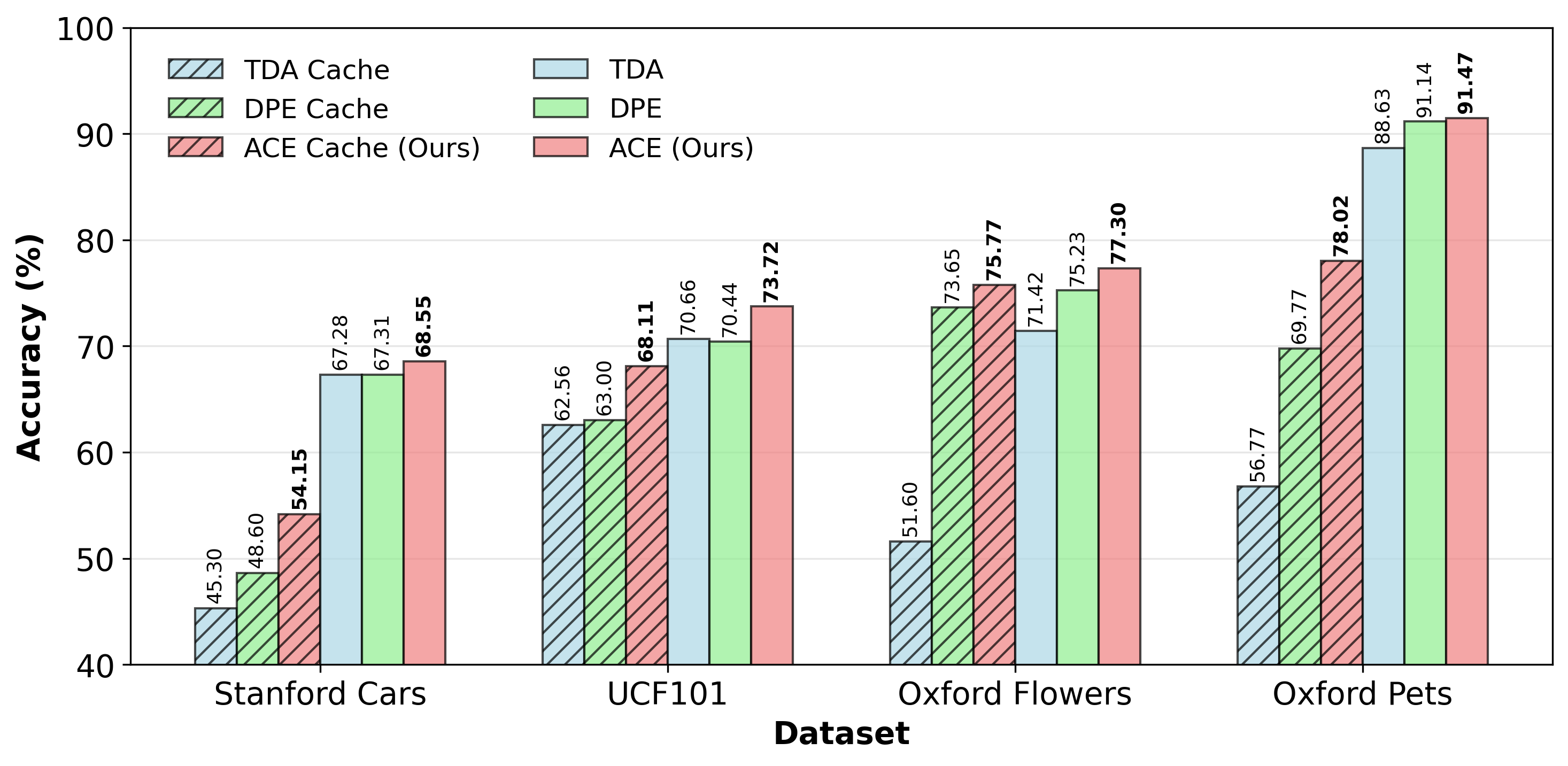}}
	\caption{Comparison of cache accuracy vs. test accuracy over four datasets between TDA~\cite{tda}, DPE~\cite{zhang2024dual}, and our ACE-Entropy.}
        \label{fig: cachereliability}
        \vspace{-10pt}
\end{figure}

Figure~\ref{fig: cachereliability} visualizes the cache accuracy and overall classification performance of TDA, DPE, and our ACE across four fine-grained datasets. ACE demonstrates exceptional results, achieving an accuracy of 68.55\% on Stanford Cars, surpassing TDA by 1.27\% and DPE by 1.24\%, and 77.30\% on Oxford Flowers, outperforming TDA by 5.88\% and DPE by 2.07\%. These gains are complemented by ACE's consistent superiority in cache accuracy across the datasets, reinforcing its ability in enhancing classification power while sustaining reliable and flexible decision boundaries via class-wise thresholding adaptation.
The synergy between these metrics underscores their complementary nature, enhancing overall performance.

\subsubsection{Efficiency Comparison}
The inference efficiency and adaptation performance of mainstream test-time adaptation (TTA) approaches are comprehensively assessed on ImageNet \cite{deng2009imagenet}, with the outcomes summarized in Table \ref{table:efficiency}.
Compared to the zero-shot CLIP \cite{radford2021learning} baseline, which achieves 59.81\% top-1 accuracy within 11 minutes, ACE attains a markedly higher accuracy of 64.41\% after 2 hours and 40 minutes of adaptation—representing an improvement of +4.60\%.
Although TPT \cite{shu2022test} and DiffTPT \cite{feng2023diverse} obtain similar accuracies of 60.74\% and 60.80\%, respectively, both require over 10 and 20 hours of computation, resulting in substantial computational overhead that constrains their practical applicability in efficiency-critical scenarios.
TDA \cite{karmanov2024efficient}, while more efficient with a runtime of 32 minutes, achieves a lower accuracy of 61.35\%.
CRG \cite{zhai2025mitigatingcachenoisetesttime} although achieves higher accuracy of 65.26\%, it significantly requires much more time to run, result in a 3 hours running.
Overall, ACE delivers the highest accuracy among all evaluated methods while maintaining a moderate adaptation time.
Its adaptive components—such as dynamic curriculum thresholding and online thresholding adaptation - appears to enable faster convergence and more robust performance.
These findings indicate that ACE strikes a more effective balance between adaptation speed and accuracy compared to alternative techniques, underscoring its methodological strengths in test-time adaptation.

\section{Conclusion}
In this paper, we present Adaptive Cache Enhancement (ACE), which decisively modifies class-wise thresholds to ensure the caching of more reliable test samples. Extensive benchmarking across a range of datasets and domains demonstrates that ACE outperforms state-of-the-art methods, particularly in the face of challenging distribution shifts. This approach is not merely practical; it is a robust solution for adapting vision-language models in dynamic environments. Our findings clearly show that effectively addressing noisy cache samples significantly enhances generalizability.
{
    \small
    \bibliographystyle{ieeenat_fullname}
    \bibliography{main}
}

\clearpage
\maketitlesupplementary

\section{Appendix}

In appendix, we provide additional details and experimental results to enhance understanding and insights into our proposed method.
This supplementary document is organized as follows:
\begin{itemize}[leftmargin=0.5cm, itemindent=0cm, itemsep=4pt,topsep=4pt,parsep=0pt]
    \item[$\bullet$] \textbf{Detailed Dataset Information:}  
    Comprehensive details about the datasets used in our experiments, including their key characteristics and distributions, are provided.
    \item[$\bullet$]  \textbf{Textual Prompts Used in Experiments:}  
    The text templates used in our experiments for each dataset are listed for reproducibility.
    \item[$\bullet$]  \textbf{Full results of ablation study:}  
    We presented further details of the ablation study on cache size and ZS cache initialization for all datasets.
    \item[$\bullet$]  \textbf{Supplementary for different seeds:}
    We provide the full performance comparison of ACE for 3 different seeds on OOD experiments.
\end{itemize}

\section{Detailed Dataset Information}
\label{subsec:dataset}

In Table~\ref{tab:dataset}, we provide a comprehensive overview of the datasets employed in our experiments, detailing key statistics such as the number of classes, the sizes of the training, validation, and testing sets, the original tasks for which each dataset was designed. This information enables a thorough understanding of the datasets' composition and their relevance to the tasks evaluated in our study. 

\begin{table}[hp]
    \resizebox{\linewidth}{!}{
    \setlength{\tabcolsep}{0.5mm}{
    \begin{tabular}{llllll}
    \toprule
Dataset                  & Classes  & Training & Validation   & Testing & Task \\ \midrule
Caltech101~\cite{fei2004learning} & 100 & 4,128 & 1,649 & 2,465& Object recognition \\
DTD~\cite{cimpoi2014describing}& 47 & 2,820 & 1,128& 1,692 &  Texture recognition\\ 
EuroSAT~\cite{helber2019eurosat}& 10 & 13,500 & 5,400& 8,100 & Satellite image recognition \\ 
FGVCAircraft~\cite{maji2013fine} & 100 & 3,334 & 3,333& 3,333 & Fine-grained aircraft recognition\\
Flowers102~\cite{nilsback2008automated} & 102 & 4,093 & 1,633& 2,463 & Fine-grained flowers recognition \\ 
Food101~\cite{bossard2014food} & 101 & 50,500& 20,200& 30,300 & Fine-grained food recognition  \\ 
ImageNet~\cite{deng2009imagenet} & 1,000 & 1.28M & -& 50,000 & Object recognition \\ 
OxfordPets~\cite{parkhi2012cats} & 37  & 2,944 & 736& 3,669 & Fine-grained pets recognition \\ 
StanfordCars~\cite{krause20133d} & 196 & 6,509 & 1,635& 8,041 & Fine-grained car recognition \\
SUN397~\cite{xiao2010sun}& 397& 15,880 & 3,970& 19,850 & Scene recognition\\ 
UCF101~\cite{soomro2012ucf101}& 101 & 7,639 & 1,898& 3,783 & Action recognition\\
\midrule
ImageNet-V2~\cite{recht2019imagenet} & 1,000 & - & -& 10,000 & Robustness of collocation  \\
ImageNet-Sketch~\cite{wang2019learning} & 1,000 & - & -&50,889 & Robustness of sketch domain\\
ImageNet-A~\cite{hendrycks2021natural}& 200 & - & -&7,500 &Robustness of adversarial attack\\
ImageNet-R~\cite{hendrycks2021many}& 200 & - & -&30,000&Robustness of multi-domains\\
    \bottomrule
    \end{tabular}
    }
    }
    \caption{\textbf{Detailed statistics of datasets used in experiments}. Note that the last 4 ImageNet variant datasets are designed for evaluation and only contain the test sets.}
    \label{tab:dataset}
\end{table}

\section{Textual Prompts Used in Experiments}
\label{sec:prompt}
In Table~\ref{tab:prompt}, we present a detailed compilation of the hand-crafted prompts specifically designed and utilized for each dataset in our experiments.

\begin{table}[!htp]
\centering
    \resizebox{\linewidth}{!}{
    \begin{tabular}{lr}
    \toprule
Dataset                  & Prompts   \\ \midrule
& ``itap of a \{\texttt{CLASS}\}.'' \\ 
ImageNet~\cite{deng2009imagenet}& ``a bad photo of the \{\texttt{CLASS}\}.'' \\ 
ImageNet-V2~\cite{recht2019imagenet}& ``a origami \{\texttt{CLASS}\}.'' \\ 
ImageNet-Sketch~\cite{wang2019learning}& ``a photo of the large \{\texttt{CLASS}\}.'' \\ 
ImageNet-A~\cite{hendrycks2021natural}& ``a \{\texttt{CLASS}\} in a video game.'' \\ 
ImageNet-R~\cite{hendrycks2021many}& ``art of the \{\texttt{CLASS}\}.'' \\ 
& ``a photo of the small \{\texttt{CLASS}\}.'' \\ \midrule
Caltech101~\cite{fei2004learning} & ``a photo of a \{\texttt{CLASS}\}.'' \\
DTD~\cite{cimpoi2014describing}& ``\{\texttt{CLASS}\} texture.'' \\ 
EuroSAT~\cite{helber2019eurosat}& ``a centered satellite photo of \{\texttt{CLASS}\}.'' \\ 
FGVCAircraft~\cite{maji2013fine} & ``a photo of a \{\texttt{CLASS}\}, a type of aircraft.'' \\
Flowers102~\cite{nilsback2008automated} & ``a photo of a \{\texttt{CLASS}\}, a type of flower.'' \\ 
Food101~\cite{bossard2014food} & ``a photo of \{\texttt{CLASS}\}, a type of food.'' \\ 
OxfordPets~\cite{parkhi2012cats} & ``a photo of a \{\texttt{CLASS}\}, a type of pet.''  \\ 
StanfordCars~\cite{krause20133d} & ``a photo of a \{\texttt{CLASS}\}.'' \\
SUN397~\cite{xiao2010sun}& ``a photo of a \{\texttt{CLASS}\}.''\\ 
UCF101~\cite{soomro2012ucf101}& ``a photo of a person doing \{\texttt{CLASS}\}.'' \\
    \bottomrule
    \end{tabular}
    }
    \caption{\looseness=-1 \textbf{Textual prompts used in experiments}. In addition to these prompts, we also employ CuPL~\cite{pratt2023does} prompts to further enhance performance.}
    \label{tab:prompt}
    \vspace{-15pt}
\end{table}

\section{Full results of ablation study}
We report full results of cross-datasets generalization settings to show the effectiveness of zero-shot statistics and choice of cache size in Table~\ref{tab:ablation}. 
\begin{table*}[t]
  \centering
  \setlength{\tabcolsep}{1.4mm}
  \begin{subtable}[t]{\textwidth}\small
    \centering
    \resizebox{\linewidth}{!}{%
    \begin{tabular}{llccccccccccc}
      \toprule
      Setting & Strat. & Aircraft & Caltech & Cars & DTD & EuroSAT & Flower & Food101 & Pets & SUN397 & UCF101 & Mean\\
      \midrule
       DPE & - & 19.80 & 90.83 & 59.26 & 50.18 & 41.67 & 67.60 & 77.83 & 85.97 & 64.23 & 61.98 & 61.93 \\ \midrule
       w/o ZS cache & Prob. & 20.97 & 90.55 & 59.63 & 52.54 & 45.94 & 69.51 & 77.47 & 87.41 & 64.92 & 61.81 & 63.08 \\
       w/ ZS cache & Prob.  & 21.33 & 90.91 & 59.84 & 55.20 & 51.02 & 70.28 & 76.83 & 87.95 & 65.46 & 61.83 & 64.07 \\ \midrule
       w/o ZS cache & Ent.  & 20.37 & 90.39 & 59.37 & 52.19 & 43.89 & 69.39 & 77.48 & 87.00 & 64.99 & 62.18 & 62.74 \\
       w/ ZS cache & Ent.  & 21.48 & 90.99 & 60.00 & 53.90 & 48.54 & 70.08 & 76.91 & 87.54 & 65.38 & 62.60 & 63.74 \\
       \midrule
       \midrule
       DPE & -  &  28.95 & 94.81 & 67.31 & 54.20 & 55.79 & 75.07 & 86.17 & 91.14 & 70.07 & 70.44 & 69.40 \\ \midrule
       w/o ZS cache & Prob.  & 29.67 & 95.06 & 68.44 & 56.21 & 55.13 & 75.80 & 86.34 & 91.21 & 70.40 & 72.06 & 70.03 \\
       w/ ZS cache & Prob.  & 29.46 & 95.13 & 68.72 & 56.38 & 54.28 & 77.47 & 85.88 & 91.47 & 70.83 & 73.62 & 70.32 \\ \midrule
       w/o ZS cache & Ent.  & 30.15 & 95.12 & 68.40 & 56.09 & 54.73 & 74.99 & 86.29 & 91.12 & 70.51 & 72.16 & 69.96 \\
       w/ ZS cache & Ent.  & 29.49 & 95.33 & 68.55 & 56.56 & 53.63 & 77.30 & 85.88 & 91.47 & 70.88 & 73.72  & 70.28 \\
      \bottomrule
    \end{tabular}%
    }
    \caption{Impact of zero-shot cache}
  \end{subtable}\hfill
  
  \begin{subtable}[t]{\textwidth}\small
    \centering
    \resizebox{\linewidth}{!}{%
    \begin{tabular}{llccccccccccc}
      \toprule
      Strat. & Cache size & Aircraft & Caltech & Cars & DTD & EuroSAT & Flower & Food101 & Pets & SUN397 & UCF101 & Mean\\
      \midrule
      Prob. & M = 3 & 20.34	& 90.63 &	57.27 &	51.30 &	41.02 &	69.96 &	76.41 &	85.94 &	63.72 &	60.32 & 61.69 \\
      Ent. & M = 3 & 19.32 &	90.91 &	57.08 &	51.12 &	42.37 &	68.17 &	76.45 &	86.02 &	63.33 &	60.69 & 61.55 \\
      \midrule
      Prob. & M = 16 & 21.33 & 90.91 & 59.84 & 55.20 & 51.02 & 70.28 & 76.83 & 87.95 & 65.46 & 61.83 & 64.07 \\
      Ent. & M = 16 & 21.48 & 90.99 & 60.00 & 53.90 & 48.54 & 70.08 & 76.91 & 87.54 & 65.38 & 62.60 & 63.74 \\
      \midrule
      \midrule
      Prob. & M = 3 & 27.69 &	95.09 &	66.26 &	55.02 &	54.14 &	77.39 &	85.50 &	89.94 &	69.38 &	71.74 & 69.22 \\
      Ent. & M = 3 & 28.68 &	95.29 &	66.04 &	54.31 &	53.32 &	77.59 &	85.50 &	90.22 &	69.30 &	73.35 & 69.36 \\
      \midrule
      Prob. & M = 16 & 29.46 & 95.13 & 68.72 & 56.38 & 54.28 & 77.47 & 85.88 & 91.47 & 70.83 & 73.62 & 70.32 \\
      Ent. & M = 16 & 29.49 & 95.33 & 68.55 & 56.56 & 53.63 & 77.30 & 85.88 & 91.47 & 70.88 & 73.72 & 70.28 \\
      \bottomrule
    \end{tabular}%
    }
    \caption{Impact of cache size}
  \end{subtable}
\caption{Ablation study on the impact of zero-shot cache initialization and cache size toward ACE.
}
\label{tab:supp_ood}
\end{table*}

\section{Supplementary for different seeds}
We report the complete results of ImageNet with 3 different seeds in Table \ref{tab:ood-main-supp}.
Since we introduce the Nnever and rarely seen adaptation, and ACE is only applied during testing, the sequence of testing samples could affect the performance.
Here, as we can see in Table \ref{tab:ood-main-supp}, the standard deviation between datasets is stable, showing that ACE performance is not affected by randomness. 

\begin{table*}[t]\small
	\renewcommand\arraystretch{1.2}
	\renewcommand{\tabcolsep}{3pt}
	\vspace{5pt}
	\centering
	\resizebox{\linewidth}{!}{
		\begin{tabular}{lccccccc}
			\toprule
			{Method}      & ImageNet   &  ImageNet-A  &  ImageNet-V2  & ImageNet-R 
			& ImageNet-S   
			&{Average}  & {OOD Average}     \\
			\midrule
			CLIP-ResNet-50~\cite{radford2021learning}       &58.16&	21.83&	51.41	&56.15	&33.37&	44.18&	40.69           \\ 
			\midrule
			Ensemble          &  59.81&	23.24&	52.91	&{60.72}	&35.48&	46.43&	43.09  \\
			CoOp~\cite{zhou2022learning}            &  63.33 &  23.06  &   55.40     &  56.60   &  34.67   &     46.61   &     42.43       \\
			\midrule
			TPT~\cite{shu2022test}          &  60.74 & 26.67  &    54.70   &  59.11   &  35.09   &    47.26    &     43.89    \\
			DiffTPT~\cite{feng2023diverse} & 60.80 & 31.06 & 55.80 & 58.80 & 37.10 & 48.71 & 45.69 \\
			TDA~\cite{karmanov2024efficient} & 61.35 & 30.29 & 55.54 & 62.58 & 38.12 & 49.58 & 46.63 \\ 
			TPS~\cite{sui2024just}	& 61.47 & 30.48 &	54.96 &	62.87 &	37.14&	49.38&	46.36 \\
			DMN-ZS~\cite{zhang2024dual}	& 63.87 &	28.57 &	56.12&	61.44 &	39.84 &	49.97 &	46.49\\ 
			DPE~\cite{dpe} & 63.41 & 30.15 & 56.72 & 63.72 & 40.03 & 50.81  & 47.66 \\
			CRG~\cite{zhai2025mitigatingcachenoisetesttime} & \textbf{65.26} & 29.69 & 56.07 & - & - & - & - \\
			
			\textbf{ACE-Probability} & \underline{64.41}$\pm$0.12 & \underline{33.50}$\pm$0.12 & \underline{57.50}$\pm$0.02 & \textbf{63.95}$\pm$0.13 & \underline{40.15}$\pm$0.01 & \underline{51.75}$\pm$0.03  & \underline{49.37}$\pm$0.02 \\
			
			\textbf{ACE-Entropy} & 64.30$\pm$0.12 & \textbf{33.55}$\pm$0.12 & \textbf{57.51}$\pm$0.03 & \underline{63.82}$\pm$0.15 & \textbf{40.25}$\pm$0.02 & \textbf{51.78}$\pm$0.04  & \textbf{49.38}$\pm$0.01 \\
			\midrule
			\midrule
			CLIP-ViT-B/16~\cite{radford2021learning}       & 66.73 &	47.87&	60.86&	73.98&	46.09&	59.11&	57.20             \\
			\midrule
			Ensemble          &  68.34&	49.89&	61.88&	{77.65}&	48.24&	61.20&	59.42  \\
			CoOp~\cite{zhou2022learning}          &  71.51 &  49.71  &   64.20     &  75.21   &  47.99   &   61.72     &   59.28  \\
			\midrule
			TPT~\cite{shu2022test}           &  68.98  & 54.77  & 63.45       &  77.06   &  47.94   &   62.44     &   60.81 \\
			DiffTPT~\cite{feng2023diverse}           &  70.30  & 55.68  & 65.10       &  75.00   &  46.80   &   62.28     &   60.52 \\
			TDA~\cite{karmanov2024efficient} & 69.51 & 60.11 & 64.67 & 80.24 & 50.54 & 65.01 & 63.89 \\ 
			TPS~\cite{sui2024just}	&70.19&	60.08&	64.73&	80.27 &	49.95&	65.04&	63.76 \\
			DMN-ZS~\cite{zhang2024dual}	& 72.25 &	58.28 &	65.17 &	78.55 &	\textbf{53.20} &	65.49 &	63.80\\
			DPE~\cite{dpe} & 71.91 & 59.63 & 65.44 & 80.40 & 52.26 & 65.93 & 64.43 \\
			CRG~\cite{zhai2025mitigatingcachenoisetesttime} & \textbf{75.01} & \textbf{63.67} & 64.66  & - & - & - & - \\
			
			\textbf{ACE-Probability} & 72.56$\pm$0.12 & \underline{63.54}$\pm$0.11 & \underline{65.86}$\pm$0.15 & \underline{81.01}$\pm$0.02 & 52.60$\pm$0.11 & \underline{67.05}$\pm$0.04 & \underline{66.01}$\pm$0.24 \\
			
			\textbf{ACE-Entropy} & \underline{72.57}$\pm$0.14 & 63.47$\pm$0.12 & \textbf{65.91}$\pm$0.16 & \textbf{81.09}$\pm$0.06 & \underline{52.66}$\pm$0.16 & \textbf{67.10}$\pm$0.11 & \textbf{66.05}$\pm$0.18 \\
			\bottomrule
		\end{tabular}
	}
	\caption{Performance comparisons on robustness to natural distribution shifts. We present top-1 accuracy (\%) results for all evaluated methods employing both ResNet-50 and ViT-B/16 visual backbones of CLIP. The best and second-best results are highlighted in \textbf{bold} and \underline{underlined}, respectively.}
	\label{tab:ood-main-supp}
	\vspace{-10pt}
\end{table*}

\begin{table*}[t]
    \footnotesize
    \centering
    {\captionsetup[subfloat]{labelfont=scriptsize}
    \subfloat[
    Effect of Curriculum Thresholding (CT) alone.  
            Reported numbers are accuracy improvements (\%) over DPE on five datasets.
    \label{tab:ct_only}
    ]
    {\begin{minipage}[t]{0.2\linewidth}{
        \begin{center}
            \begin{tabular}{l c}
            \toprule
            \textbf{Dataset} & \textbf{$\Delta$Acc (\%)} \\
            \midrule
            Caltech & +0.20 \\
            DTD        & +1.10 \\
            Flowers & +0.80 \\
            Pets & +0.1 \\
            UCF101     & +1.2 \\
            \midrule
            \textbf{Average} & \textbf{+0.68} \\
            \bottomrule
            \end{tabular}
        \end{center}
    }\end{minipage}
    }
    }
    \qquad
    {\captionsetup[subfloat]{labelfont=scriptsize}
    \subfloat[
    Effect of EMA-based refinement on top of CT.  
            Accuracy improvements are relative to DPE.
    \label{tab:ema_only}
    ]
    {\begin{minipage}[t]{0.2\linewidth}{
        \begin{center}
            \begin{tabular}{l c}
            \toprule
            \textbf{Dataset} & \textbf{$\Delta$Acc (\%)} \\
            \midrule
            Caltech&	+0.30\\
            DTD&	+1.70\\
            Flower&	+1.50\\
            Pets&	+0.70\\
            UCF&	+1.40\\
            \midrule
            \textbf{Average} & \textbf{+1.12} \\
            \bottomrule
            \end{tabular}
        \end{center}
    }\end{minipage}
    }
    }
    \qquad
    {\captionsetup[subfloat]{labelfont=scriptsize}
    \subfloat[
    Effect of the exploration mechanism on top of CT + EMA. 
            Accuracy improvements are relative to DPE.
    \label{tab:exploration}
    ]
    {\begin{minipage}[t]{0.2\linewidth}{
        \begin{center}
            \begin{tabular}{l c}
            \toprule
            \textbf{Dataset} & \textbf{$\Delta$Acc (\%)} \\
            \midrule
            Caltech &	+0.45\\
            DTD &	+2.10\\
            Flower &	+2.30\\
            Pets&	+1.50\\
            UCF&	+2.80\\
            \midrule
            \textbf{Average} & \textbf{+1.83} \\
            \bottomrule
            \end{tabular}
        \end{center}
    }\end{minipage}
    }
    }
    \qquad
    {\captionsetup[subfloat]{labelfont=scriptsize}
    \subfloat[
    Impact of zero-shot initialization of class-wise thresholds added to all components.  
    Accuracy improvements relative to DPE baseline.
    \label{tab:zs_init}
    ]
    {\begin{minipage}[t]{0.2\linewidth}{
        \begin{center}
            \begin{tabular}{l c}
            \toprule
            \textbf{Dataset} & \textbf{$\Delta$Acc (\%)} \\
            \midrule
            Caltech &	+0.32\\
            DTD &	+2.18\\
            Flower &	+2.40\\
            Pets &	+0.33\\
            UCF &	+3.18\\
            \midrule
            \textbf{Average} & \textbf{+1.68} \\
            \bottomrule
            \end{tabular}
        \end{center}
    }\end{minipage}
    }
    }
    \vspace{-10pt}
    \caption{Ablation study of each of the thresholding components.}
    \label{tab:table_thresh}
\end{table*}

\section{Ablation of Thresholding Components}

To better understand the contribution of each part of our adaptive thresholding
design, we decompose ACE into four sub-components corresponding to the
individual tables in Fig.~\ref{tab:table_thresh}. Each sub-table (a)--(d)
reports accuracy improvements (\%) over the DPE baseline across the five
datasets used in our main evaluation.

\paragraph{(a) Curriculum Thresholding (CT).}
Table~\ref{tab:ct_only} isolates the effect of CT alone.  
By allowing class-wise thresholds to evolve based on the model's
progress, CT already provides substantial improvements over DPE across all
datasets, most notably on DTD and Flowers102.  
These results demonstrate that dynamically adjusting thresholds is fundamental
to increasing cache purity and stabilizing early adaptation.

\paragraph{(b) EMA-Based Refinement.}
Adding EMA smoothing on top of CT (Table~\ref{tab:ema_only}) further improves
accuracy on every dataset.  
EMA suppresses noisy fluctuations in threshold values and stabilizes
pseudo-label evolution, leading to a consistent gain of roughly +1\% over
CT alone.  
This confirms that temporal smoothing is crucial for reliable threshold
updates in an online setting.

\paragraph{(c) Exploration for Rare Classes.}
Table~\ref{tab:exploration} evaluates the effect of enabling our exploration
mechanism.  
By decreasing thresholds for underrepresented classes, exploration increases
cache diversity and improves performance on datasets with high intra-class
variance or long-tail structure.  
The gains over (b) are the largest among all incremental components,
highlighting exploration as a key driver for handling rare or difficult
categories.

\paragraph{(d) Zero-Shot Initialization (ZS-Init).}
Finally, Table~\ref{tab:zs_init} shows the effect of adding zero-shot
threshold initialization.  
ZS-Init provides a strong, distribution-aware starting point for each class,
preventing early miscalibration and significantly improving early adaptation
stability.  
When combined with CT, EMA, and exploration, ZS-Init produces the highest
overall accuracy improvement, confirming its role in accelerating convergence
and improving cache reliability.

\paragraph{Summary.}
Across all four sub-components, each contributes positively and
incrementally to the final performance of ACE.  
The progression from (a) to (d) demonstrates that curriculum scheduling,
temporal smoothing, exploration, and zero-shot initialization are
individually beneficial and complementary when combined.

\end{document}